\documentclass[10pt,twocolumn,letterpaper]{article}

\usepackage{cvpr}
\usepackage{multirow}
\usepackage{algorithmic}
\usepackage{algorithm}
\usepackage[algo2e]{algorithm2e}
\usepackage{booktabs}
\usepackage{pifont}
\usepackage{placeins}
\newcommand{\cmark}{\ding{51}}% check mark
\newcommand{\xmark}{\ding{55}}% x mark
\usepackage{booktabs}    % \toprule, \midrule, \bottomrule
\usepackage{multirow}    % multi-row 셀 병합
\usepackage{multicol}    % multi-column 병합 (있으면 편함)
\usepackage{makecell}    % 셀 내 줄바꿈, 정렬 조정 가능
\usepackage{array}       % 열 정렬 옵션 강화
\usepackage{adjustbox}   % \resizebox{width}{!}{} 지원
\usepackage{graphicx}    % 그림 삽입 및 \resizebox
\usepackage{rotating}    % \rotatebox (세로 텍스트용)
\usepackage{caption}
\definecolor{cvprblue}{rgb}{0.21,0.49,0.74}
\usepackage[pagebackref,breaklinks,colorlinks,allcolors=cvprblue]{hyperref}

\def\confName{CVPR}
\def\confYear{2026}

\title{Generalizable Human Gaussian Splatting
via Multi-view Semantic Consistency}

\author{
    Jingi Kim \quad Wonjun Kim\thanks{Corresponding author} \\ 
    Konkuk University\\
    {\tt\small \{jingi0614, wonjkim\}@konkuk.ac.kr}
}
\begin{document}
\maketitle
\begin{abstract}
\quad Recently, generalizable human Gaussian splatting from sparse-view inputs has been actively studied for the photorealistic human rendering.
Most existing methods rely on explicit geometric constraints or predefined structural representations to accurately position 3D Gaussians.
Although these approaches have shown the remarkable progress in this field, they still suffer from inconsistent feature representations across multi-view inputs due to complex articulations of the human body and limited overlaps between different views.
To address this problem, we propose a novel method to accurately localize 3D Gaussians and ultimately improve the quality of human rendering.
The key idea is to unproject latent embeddings encoded from each viewpoint into a shared 3D space through predicted depth maps and recalibrate them belonging to the same body part based on cross-view attention.
This helps the model resolve the spatial ambiguity occurring in highly textured regions as well as occluded body parts, thus leading to the accurate localization of 3D Gaussians. 
Experimental results on benchmark datasets show that the proposed method efficiently improves the performance of generalizable human Gaussian splatting from sparse-view inputs.
The code and model are publicly available at link\footnote{\href{https://github.com/DCVL-3D/GHGS-MVSC_release}{\nolinkurl{https://github.com/DCVL-3D/GHGS-MVSC_release}}}.
% The code and model are publicly available at {\href{https://github.com/DCVL-3D/GHGS-MVSC_release}{\nolinkurl{https://github.com/DCVL-3D/GHGS-MVSC_release}}}.
\end{abstract}    
\vspace{-3.5mm}
\section{Introduction}
\vspace{-1.5mm}
% Paragraph 1
\quad In the field of novel view synthesis (NVS), 3D Gaussian splatting (3DGS)~\cite{3DGS} has become mainstream, which explicitly models a scene with a set of 3D Gaussian primitives and renders it through a differentiable rasterizer in real-time.
Recently, many studies~\cite{gaussianavatar, kocabas2024hugs, lei2024gart, wen2024gomavatar, Qian20243DGS-Avatar} have applied this 3DGS scheme to resolve the problem of human-specific rendering by leveraging various regularization techniques with the human parametric model (e.g., SMPL~\cite{loper2015smpl}).
Although these approaches have achieved the significant progress, the tedious per-subject optimization and the dependence on dense-view inputs still limit their practicality in real-world scenarios.

%%% Fig1
\begin{figure}[t]  % [t]: top 위치
    \centering
    \includegraphics[width=\linewidth]{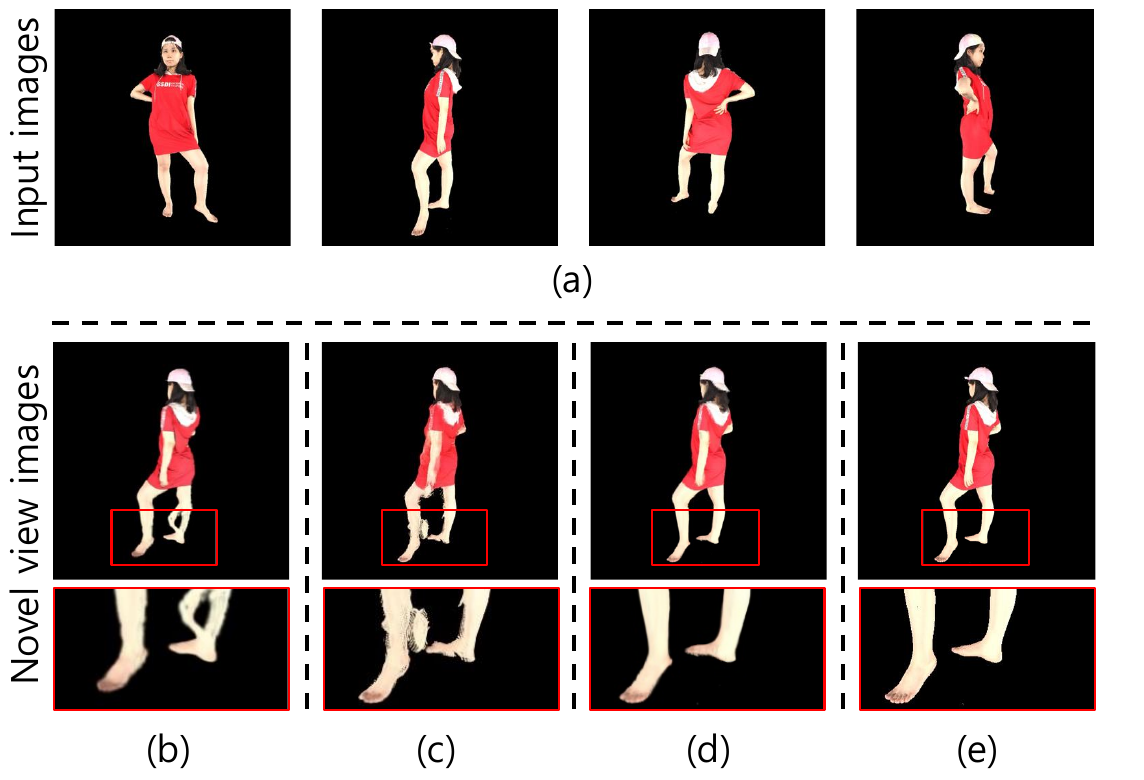}
    % \vspace{-7mm}
    \caption{Examples of generalizable human Gaussian splatting. (a) Input images (four-view inputs). (b) Result by~\cite{gps-gaussian}. (c) Result by~\cite{rogsplat}. (d) Result by the proposed method. (e) Ground truth.}
    \label{fig:fig1}
\end{figure}
%%% Paragraph 2

To alleviate these limitations, recent studies have explored generalizable approaches that estimate parameters of 3D Gaussians for unseen human subjects from sparse-view inputs in a feed-forward manner.
Specifically, there have been several approaches~\cite{ghg, lifegom, rogsplat} to utilize UV maps or meshes, which are computed from the SMPL model, for initializing and refining the position of Gaussians.
However, Gaussians are inaccurately positioned on textured surfaces such as hair and loose clothing due to the skinned property of the SMPL model.
Moreover, the rendering quality is highly dependent on the estimation accuracy of SMPL parameters, which is often unreliable when dynamic motions or self-occlusions occur.
On the other hand, the depth information, which is estimated from explicit geometric constraints, e.g., plane-sweep cost volume~\cite{gps-gaussian}, epipolar geometry~\cite{hu2024eva}, etc., has been adopted to localize 3D Gaussians.
Point clouds unprojected from pixel-wise depth values make Gaussians positioned flexibly as well as accurately in the 3D space, however, the insufficient overlap between sparse-view inputs probably yields mismatched correspondences and ultimately impedes the accurate depth estimation.
Consequently, the aforementioned challenges make it difficult to localize 3D Gaussians precisely onto the physical surface, which leads to the performance drop of rendering results, as shown in Figs.~\ref{fig:fig1}(b) and (c).

In this paper, we propose a novel generalizable human Gaussian splatting method from sparse-view inputs. 
The key idea is to recalibrate latent embeddings encoded from each viewpoint by utilizing the semantic consistency in the 3D space for accurately localizing 3D Gaussians. 
Specifically, latent embeddings encoded from each viewpoint are first unprojected into the shared 3D space through predicted depth maps. 
Subsequently, latent embeddings, which are spatially close in this 3D space, are used to determine whether they belong to the same body part or not. 
To do this, the semantic consistency is computed based on semantic features (e.g., DINO features~\cite{oquabdinov2}) corresponding to latent embeddings, which are extracted from the intermediate step of the encoding process. 
Finally, latent embeddings belonging to the same body part are recalibrated via cross-view attention to give the part-attentive information for estimating Gaussian attributes. 
This helps the model alleviate the spatial ambiguity occurring in highly textured regions as well as occluded body parts, thus leading to the accurate localization of 3D Gaussians.
The main contribution of the proposed method can be summarized as follows:
\begin{itemize}
    \item 
    We propose to make latent embeddings provide the part-attentive information based on the semantic consistency, which helps the model accurately estimate Gaussian attributes. By computing cross-view attention on semantically consistent embeddings across multi-view inputs, i.e., embeddings belonging to the same body part, 3D Gaussians can be accurately localized via such recalibrated results, thus leading to the significant improvement of rendering results.

    \item 
We propose to unproject the latent embedding itself into the 3D space. By doing this, the spatial relation between multi-view inputs can be accurately established in the 3D space, which improves the performance of the depth decoder. This also helps aggregate latent embeddings belonging to the same body part in the 3D space.   
\end{itemize}

\section{Related Works}
\label{sec:Related Works}
\quad In this Section, we provide a brief review of recent methods for human rendering. These methods can be divided into two main groups based on 3D human representation: neural radiance fields (NeRF)~\cite{Nerf} and  3D Gaussian splatting (3DGS)~\cite{3DGS}.

\subsection{NeRF-based Methods for Human Rendering}
\quad In the last few years, NeRF~\cite{Nerf} has significantly improved the performance of NVS, which encodes a given 3D scene into implicit radiance fields by mapping 3D coordinates and the view direction to color and density values through a simple neural network.
Building upon these promising results, various studies have applied this NeRF scheme to reconstruct human subjects from sparse-view inputs or monocular videos, enabling applications such as the avatar animation~\cite{li2022tava, wang2022arah, liu2021neural} and the content editing~\cite{park2021nerfies, chen2024meshavatar, chen2022relighting4d, park2021hypernerf}.
Despite this advance, NeRF-based approaches typically require the per-subject optimization, which fails to generalize to unseen subjects. 
To alleviate this limitation, generalizable approaches have been explored by conditioning the radiance field on image-guided features~\cite{yu21pixelnerf, su2021nerf, wang2021ibrnet}, while leveraging the human parametric model~\cite{loper2015smpl,kwon2021neural,gao2022mps, hu2023sherf} or 3D skeleton keypoints~\cite{mihajlovic2022keypointnerf} for structural guidance.
However, these approaches remain computationally expensive due to the volumetric rendering process.
Although considerable efforts have been made to accelerate NeRF-based approaches~\cite{fridovich22plenoxels, muller22instant, chen2022tensorf, fridovich2023kplanes}, developing NeRF-based human rendering methods capable of real-time operation remains a challenge.

\begin{figure*}[t]
    \centering
    \includegraphics[width=\textwidth]{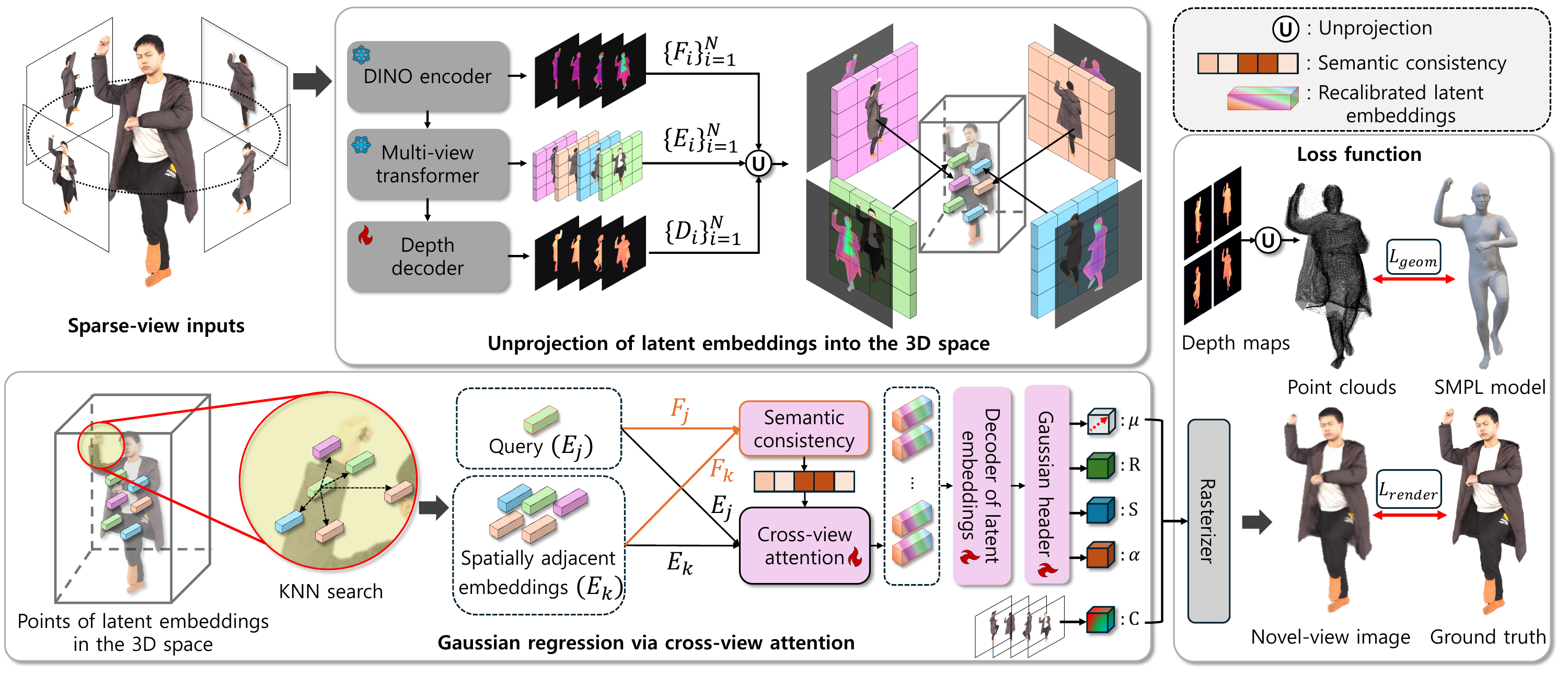}
    \vspace{-5mm}
    \caption{
        Overall architecture of the proposed method. The VGGT~\cite{wang2025vggt} encoder is used to extract semantic features and latent embeddings, which are unprojected into the shared 3D space by using predicted depth maps. Within the 3D space, spatially adjacent embeddings are recalibrated by applying cross-view attention based on the semantic consistency. Recalibrated points of latent embedding are fed into the Gaussian header to estimate Gaussian attributes.
    }
    \label{fig:overview}
\end{figure*}

\subsection{3DGS-based Methods for Human Rendering}
\quad Recently, 3DGS~\cite{3DGS} has emerged in the field of NVS, which represents a given scene as a set of 3D Gaussians and renders the image through a differentiable rasterization technique.
By replacing neural network-based representations with point-based 3D Gaussians, 3DGS dramatically improves the rendering efficiency, enabling real-time applications of 3DGS.
Inspired by the promising performance of 3DGS, various follow-up studies have been introduced in the field of 3D human reconstruction, however, these approaches typically require separate optimization for each subject, which makes generalization to unseen humans difficult.
To cope with this problem, several methods have introduced pixel-wise Gaussian parameter maps by incorporating the depth estimation based on explicit geometric constraints.
For example, Zheng \textit{et al.}~\cite{gps-gaussian} utilized plane-sweep cost volume for stereo matching. Hu \textit{et al.}~\cite{hu2024eva} leveraged 1D-window attention based on epipolar geometry to improve the rendering quality.
Meanwhile, several methods have attempted to achieve generalization by initializing the position of Gaussians through the SMPL model and refining the localization with learnable offsets.
Specifically, Kwon \textit{et al.}~\cite{ghg} regressed 3D Gaussian parameters on the 2D UV space of a human template and introduced a multi-scaffold representation to bridge geometric gaps between the skinned human body and real human shapes.
Wen \textit{et al.}~\cite{lifegom} refined SMPL vertices in an iterative framework and attached 3D Gaussians to the refined mesh surface.
Most recently, Xiao \textit{et al.}~\cite{rogsplat} constructed a set of coarse Gaussians by refining SMPL vertices with pixel-level and voxel-level features, and further predicted local offsets to obtain fine Gaussians that represent detailed geometry of the target subject.
Even though such methods have brought the significant progress in human rendering, they still suffer from insufficient visual cues and complicated articulations of the human body.
% Even though such methods have brought significant progress in human rendering, they still suffer from insufficient visual cues and complicated articulations of the human body.
\section{Proposed Method}

\subsection{Preliminaries}
\quad 3D Gaussian Splatting (3DGS)~\cite{3DGS} represents a scene using a set of 3D Gaussian primitives and synthesizes images via a differentiable rasterization process. 
Each Gaussian is defined by its mean position $\boldsymbol{\mu}_i \in \mathbb{R}^3$ and covariance matrix $\boldsymbol{\Sigma}_i \in \mathbb{R}^{3 \times 3}$, formulated as:
\begin{equation}
G_i(\boldsymbol{x}) = \exp\left( -\frac{1}{2} (\boldsymbol{x} - \boldsymbol{\mu}_i)^T \boldsymbol{\Sigma}_i^{-1} (\boldsymbol{x} - \boldsymbol{\mu}_i) \right),
\label{eq:1}
\end{equation}
where $\boldsymbol{x}$ denotes a 3D position.
The covariance $\boldsymbol{\Sigma}_i$ is further decomposed into a rotation matrix $\mathbf{R}_i \in \mathbb{R}^{3 \times 3}$ and a diagonal scaling matrix $\mathbf{S}_i\in\mathbb{R}^{3 \times 3}$ as $\boldsymbol{\Sigma}_i = \mathbf{R}_i \mathbf{S}_i \mathbf{S}_i^{T} \mathbf{R}_i^{T}$.
In addition, each Gaussian is associated with an opacity coefficient $o_i\in \mathbb{R}^{1}$ and $K$-dimensional spherical harmonic coefficients $\mathbf{f}_i\in \mathbb{R}^{{(k+1)}^2}$, which represent view-dependent color components.
The final pixel color $C(\mathbf{p})$ is obtained by alpha-blending, which can be formulated as follows:
\begin{equation}
C(\mathbf{p}) = \sum_{i=1}^{N} c_i \, \alpha_i \, \prod_{j=1}^{i-1} (1 - \alpha_j),
\label{eq:color}
\end{equation}
where $c_i$ is the color decoded from $\mathbf{f}_i$, and $\alpha_i = o_i \cdot G_i^{2D}$ denotes the per-pixel opacity determined by the 2D projection $G_i^{2D}$ of the $i$-th Gaussian, respectively.

%%%%%%%%%%%%%%%%%%%%%%%%%%%%%%%%%%%%%%%%%%%%%%%%%%%%%%

\subsection{Unprojection of Latent Embeddings into the 3D Space}

\quad Since spatial relationships between multi-view inputs cannot be accurately established in the 2D image domain, we propose to unproject latent embeddings into the 3D space using the predicted depth information.
Specifically, we adopt the pre-trained VGGT encoder~\cite{wang2025vggt} as our backbone to encode multi-view inputs into latent embeddings.
VGGT encoder consists of a self-supervised vision transformer (i.e., DINO~\cite{oquabdinov2}) and a multi-view transformer based on alternating attention, which enables the model to learn correspondence between multi-view inputs without explicit geometric constraints.
Owing to this design choice, we are able to extract semantic features, i.e., DINO features, without additional learning, as well as latent embeddings driven by the multi-view transformer, which capture spatial contexts and geometric relations across multi-view inputs.
Subsequently, we predict the per-view depth map estimated from latent embeddings by using a DPT-style decoder~\cite{ranftl2021vision}.

Once the per-view depth map is given, semantic features and latent embeddings, which are extracted from each viewpoint, are unprojected into the shared 3D space using the corresponding depth values and calibrated camera parameters as introduced in~\cite{jain2024odin}. 
This process converts 2D grids into spatially aligned 3D points. We denote such unprojected results of semantic features and latent embeddings as $F\in \mathbb{R}^{N_P \times C}$ and $E \in \mathbb{R}^{N_P \times C}$, where $N_P$ and $C$ are the number of points and channels. 
These 3D representations provide explicit 
spatial coordinates that are subsequently used for recalibration of latent embedding via semantic consistency.

\begin{figure}[t]  % [t]: top 위치
    \centering
    \includegraphics[width=\linewidth]{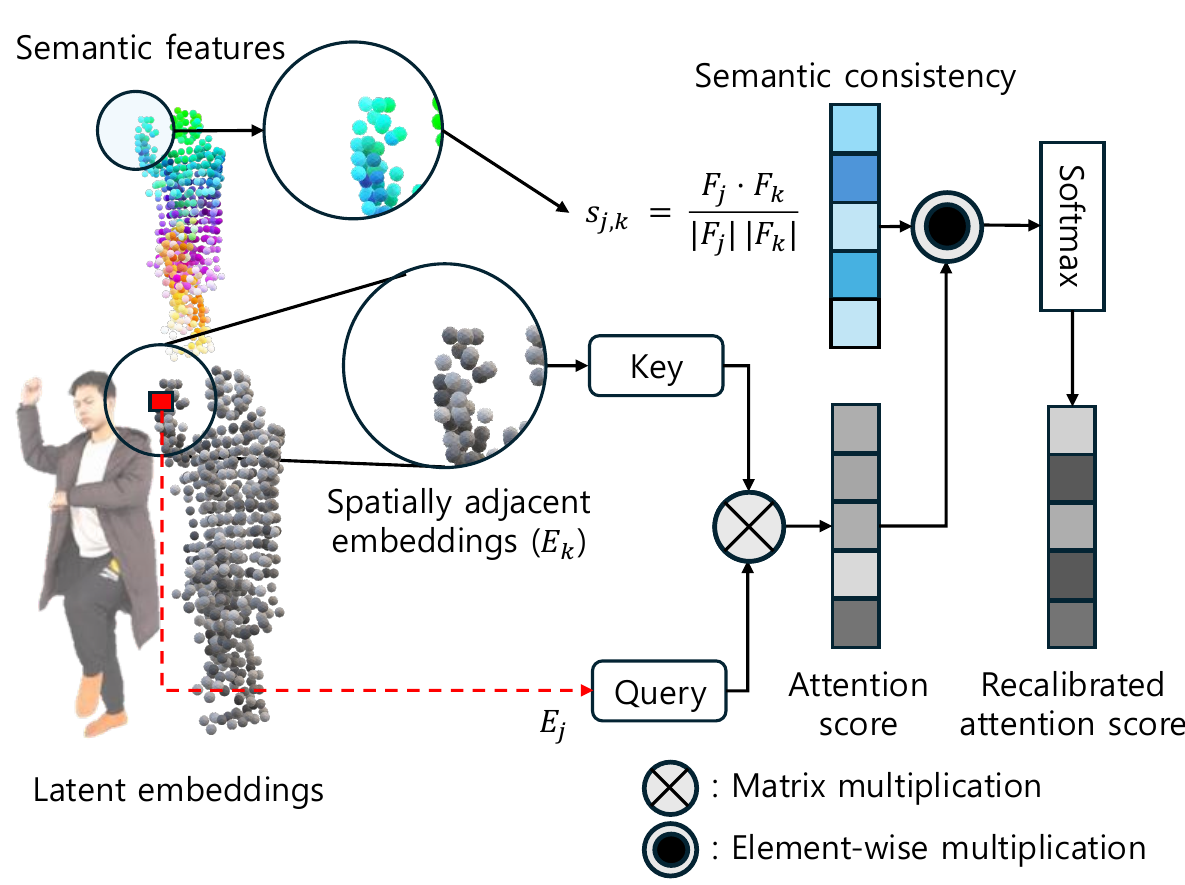}
    \caption{An example of cross-view attention to recalibrate latent embeddings through semantic consistency.}
    \label{fig:fig3}
\end{figure}

%%% Fig
\begin{figure*}[t]
    \centering
    \includegraphics[width=\textwidth]{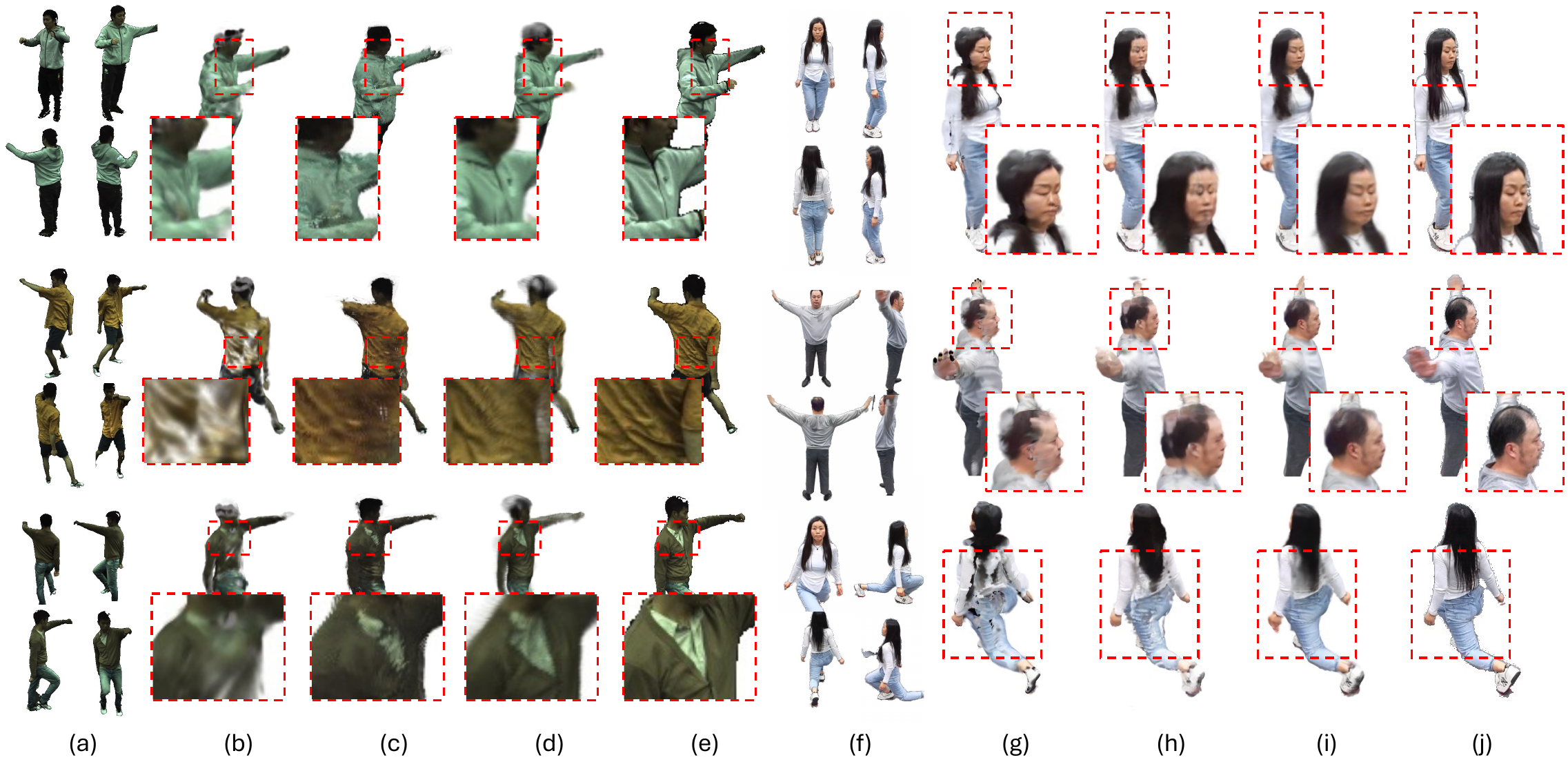}
    \vspace{-6mm}
    
    \caption{
     Results of novel view synthesis via generalizable human Gaussian splatting on ZJU-Mocap~\cite{kwon2021neural} and HuMMan~\cite{cai2022humman} datasets. (a, f) Input images. (b, g) Results by GHG~\cite{ghg}. (c, h) Results by RoGSplat~\cite{rogsplat}. (d, i) Results by the proposed method. (e, j) Ground truth.
    }
    \label{fig:zju, humman}
\end{figure*}

% \vspace{3mm}
\begin{table*}[t]
\small
\centering
\fontsize{9.5pt}{11pt}\selectfont
\renewcommand{\arraystretch}{1.1}
\setlength{\tabcolsep}{6pt}
\begin{tabular}{l  c c c | c c c}
\toprule
{\multirow{2}{*}[-0.5ex]{Methods}}
& \multicolumn{3}{c}{ZJU-Mocap} 
& \multicolumn{3}{c}{HuMMan} \\
\cmidrule(lr){2-4} \cmidrule(lr){5-7}
& PSNR($\uparrow$) & SSIM($\uparrow$) & LPIPS($\downarrow$)
& PSNR($\uparrow$) & SSIM($\uparrow$) & LPIPS($\downarrow$) \\
\midrule
% \midrule
\toprule
GHG~\cite{ghg}           
& 27.94 & 0.9421 & 0.0540
& 22.40 & 0.8915 & 0.0945 \\

RoGSplat~\cite{rogsplat} 
& 30.12 & 0.9613 & \textbf{0.0459}
& 24.94 & 0.9390 & \textbf{0.0683} \\

\textbf{Proposed method}         
& \textbf{30.58} & \textbf{0.9621} & 0.0463
& \textbf{25.06} & \textbf{0.9392} & 0.0690 \\
\bottomrule
\end{tabular}
\caption{
\label{table:zju_humman}
Quantitative comparison on ZJU-Mocap and HuMMan datasets. 
The best results are shown in bold.
}
\end{table*}

\subsection{Gaussian Regression via Cross-view Attention}

\quad After the unprojection step, all the grids of the latent embedding are represented based on 3D points in the shared 3D space as shown in Fig.~\ref{fig:overview}, where spatial relationships between multi-view inputs can be explicitly established. 
However, due to inevitable depth errors, latent embeddings corresponding to the same body part may still be spatially misaligned. 
To alleviate this limitation, latent embeddings are recalibrated via cross-view attention. 
This scheme enables latent embeddings to have part-attentive representations, which gives a great help for Gaussian regression.
Specifically, points of latent embeddings, which are spatially close, are first identified through the K-nearest neighbor (KNN) search in the 3D space. 
% Let $j$ and $k$ denote the indices of latent embeddings sampled from different views, where $j$ represents the query embedding and $k \in \mathcal{N}(j)$ refers to its spatially neighboring embeddings obtained by KNN.
Note that we index points of latent embeddings as $j$. For each query index $j$, we define $\mathcal{N}(j)$ as the set of its spatial neighbors obtained by KNN.
Since points of latent embeddings, which are elements of the set $\mathcal{N}(j)$, probably belong to the same body part, thus it is thought that those are semantically consistent.
By applying cross-view attention to this set, such embedding points are recalibrated to consider the part-attentive information.
Based on recalibrated latent embeddings, 3D Gaussians can be localized more accurately, thus the quality of human rendering is ultimately improved. 
Formally, the attention weight between the $j$-th and $k$-th points of latent embeddings is computed as follows:
% \begin{equation}
% % \alpha_{j,k} = 
% % \mathrm{Softmax}\left(
% % \frac{(W_{query} E_j)^\top (W_{key} E_k)}{\sqrt{d}}
% % \cdot s_{j,k}
% % \right),
% % \label{eq:semantic_attention}
% % \end{equation}
\begin{equation}
\alpha_{j,k} = 
\mathrm{Softmax}\left(
\frac{(W_{query} E_j) (W_{key} E_k)^\top}{\sqrt{d}}
\cdot s_{j,k}
\right),
\label{eq:semantic_attention}
\end{equation}
where $W_{query}$, $W_{key}$, and $W_{value}$ are learnable projection matrices for query, key, and value, respectively.
$\sqrt{d}$ is a normalization factor for the attention score.
$s_{j,k}$ denotes the semantic consistency, which can be computed by using the cosine similarity between two points of semantic features as follows:
\begin{equation}
s_{j,k} = \frac{F_j \cdot F_k}{\|F_j\| \, \|F_k\|}.
\end{equation}
It is noteworthy that points of latent embedding on the same body part are effectively emphasized by multiplying the semantic consistency $s_{j,k}$ with the attention score.
The recalibrated embedding $\tilde{E}_j$ is obtained as follows:
\begin{equation}
\tilde{E_j} = 
\sum_{k \in \mathcal{N}(j)} \alpha_{j,k} \, W_{value} E_k.
\end{equation}
Through this recalibration process based on the semantic consistency, the spatial ambiguity occurring in highly textured regions as well as occluded body parts is reduced, which helps the model accurately estimate Gaussian attributes.  
The detailed process of recalibrating points of latent embeddings via multi-view semantic consistency is shown in Fig.~\ref{fig:fig3}.

To obtain per-pixel Gaussian attributes, recalibrated points of latent embeddings are projected back onto the 2D grid and fed into a DPT-style decoder~\cite{ranftl2021vision}. 
The decoder progressively upsamples the projected 2D grid to the full resolution of the input image and yields Gaussian descriptors at each pixel location.
Following the design of prior works~\cite{gps-gaussian, rogsplat}, the decoder is followed by multiple parameter-specific heads that predict attributes of 3D Gaussians (i.e., position offset, rotation, scaling, and opacity) with appropriate activation functions to ensure the physical validity.
The color attribute is directly inherited from the corresponding RGB values of the input images, which avoids redundant regression and preserves photometric consistency across multi-view inputs~\cite {gps-gaussian}.
Through this decoding process, the semantically recalibrated embeddings are transformed into dense and geometrically consistent Gaussian primitives.
Finally, such 3D Gaussian primitives are rendered using a rasterizer-based Gaussian splatting pipeline~\cite{3DGS} to produce the final human appearance.

\subsection{Loss Function}
\quad The model is trained end-to-end with a composite loss function $\mathcal{L}_{total}$ that combines a photometric rendering objective and a geometric regularization term:
\begin{equation}
    \mathcal{L}_{total} = \mathcal{L}_{render} + \lambda_{geom}\mathcal{L}_{geom},
\end{equation}
where $\lambda_{geom}$ controls the weight of the geometric term, which is set to $1$.
The rendering loss $\mathcal{L}_{render}$ supervises the photorealistic quality of synthesized views by measuring the discrepancy between the rendered image $\hat{I}$ and the ground-truth image $I_{gt}$, given as~\cite{3DGS}:
\begin{equation}
    \mathcal{L}_{render} = \lambda_{L_1}||\hat{I}-I_{gt}||_{1} + \lambda_{SSIM}(1-\text{SSIM}(\hat{I},I_{gt})).
\end{equation}
This combination of $L_1$ and SSIM terms balances structural fidelity and fine-grained appearance consistency. Note that $\lambda_{L_1}$ and $\lambda_{SSIM}$ are empirically set to $0.8$ and $0.2$, respectively.
The auxiliary geometric loss $\mathcal{L}_{geom}$ regularizes the initial result of depth estimation under sparse-view ambiguity.  
It softly constrains the point cloud $\mathcal{P}_{all}$, which is unprojected from predicted depth maps, to conform to the overall human shape.  
This is implemented by using the Chamfer distance (CD)~\cite{fan2017point} between $\mathcal{P}_{all}$ and the ground truth of SMPL vertices $\mathcal{V}_{smpl}$ as follows:
\begin{equation}
    \mathcal{L}_{geom} = CD(\mathcal{P}_{all}, \mathcal{V}_{smpl}).
\end{equation}
This term acts only as a weak geometric prior that stabilizes the depth predictor,  
while the final 3D Gaussian localization is primarily guided by the rendering objective and the recalibration process via multi-view semantic consistency.

\begin{figure*}[t]
    \centering
    \includegraphics[width=\textwidth]{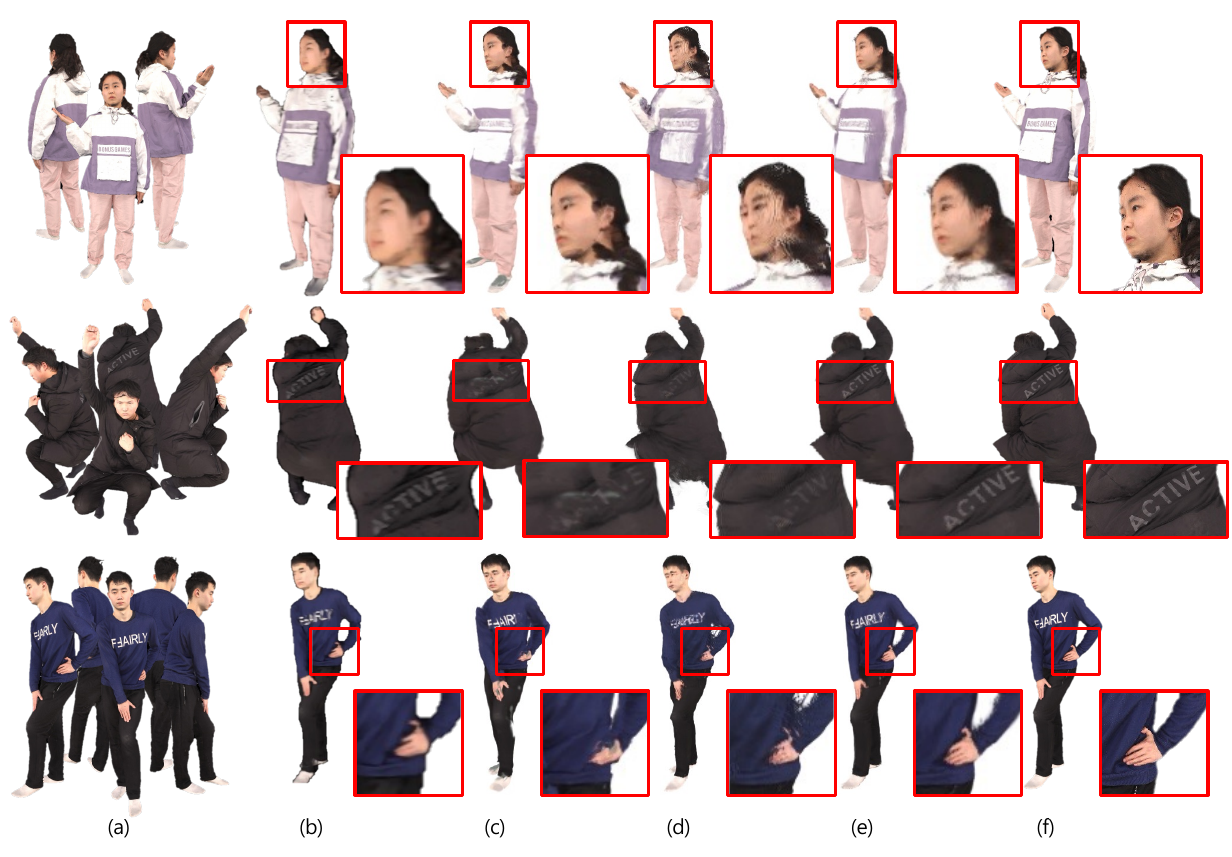}
    % \vspace{-0.8cm}
    \caption{
     Results of novel view synthesis via generalizable human Gaussian splatting on the THuman2.0~\cite{yu2021function4d} dataset. (a) Input images. (b) Results by GPS-Gaussian~\cite{gps-gaussian}. (c) Results
by GHG~\cite{ghg}. (d) Results by RoGSplat~\cite{rogsplat}. (e) Results by the proposed method. (f) Ground truth.
    }
    \label{fig:thuman}
\end{figure*}

\vspace{-5mm}

\begin{table*}[t]
\centering
\resizebox{\textwidth}{!}{
\begin{tabular}{c l| c c c | c c c | c c c}
\hline
\toprule
\multicolumn{2}{c}{\multirow{2}{*}[-0.5ex]{Methods}} & \multicolumn{3}{c}{3-view} & \multicolumn{3}{c}{4-view} & \multicolumn{3}{c}{5-view}\\ 
\cmidrule(lr){3-5}\cmidrule(lr){6-8}\cmidrule(lr){9-11}
\multicolumn{2}{c}{} & \multicolumn{1}{c}{PSNR($\uparrow$)} & \multicolumn{1}{c}{SSIM($\uparrow$)} & \multicolumn{1}{c}{LPIPS($\downarrow$)} & 
\multicolumn{1}{c}{PSNR($\uparrow$)} & \multicolumn{1}{c}{SSIM($\uparrow$)} & \multicolumn{1}{c}{LPIPS($\downarrow$)} & 
\multicolumn{1}{c}{PSNR($\uparrow$)} & \multicolumn{1}{c}{SSIM($\uparrow$)} & \multicolumn{1}{c}{LPIPS($\downarrow$)} \\ 
\midrule
\toprule
\multirow{5}{*}[0ex]
& GPS-Gaussian~\cite{gps-gaussian} & - & - & - & - & - & - & 26.54 & 0.9539 & 0.0610\\
& EVA-Gaussian~\cite{hu2024eva} & - & - & - & 26.31 & 0.9555 & 0.0391 & 27.54 & 0.9614 & 0.0297\\
& GHG~\cite{ghg} & 24.84 & 0.9232 & 0.0719 & 25.84 & 0.9462 & 0.0633 & 26.53 & 0.9571 & 0.0398\\
& RoGSplat~\cite{rogsplat} & 26.32 & 0.9478 & 0.0530 & 28.94 & 0.9615 & 0.0433 & 30.98 & 0.9711 & 0.0341\\
& \textbf{Proposed method} & \textbf{27.81} & \textbf{0.9521} & \textbf{0.0517} & 
\textbf{30.93} & \textbf{0.9710} & \textbf{0.0334} & 
\textbf{31.54} & \textbf{0.9729} & \textbf{0.0269}\\
\bottomrule
\hline
\end{tabular}}
\caption{
Performance comparison of generalizable human Gaussian splatting
on the THuman2.0~\cite{yu2021function4d} dataset.
The best is shown in bold.
}
\label{table:thuman}
\end{table*}

\section{Experimental Results}

\subsection{Implementation Details}
\quad The proposed method is implemented on the PyTorch~\cite{paszke2017automatic} framework. All the experiments were performed on a single PC with an Intel i9-10980XE@3.6 GHz CPU and NVIDIA GeForce RTX 3090 GPU. 
All the parameters of the proposed network are updated by the AdamW optimizer, where momentum factors are set to 0.9 and 0.999, respectively.
We use the batch size of 1 and the learning rate is set to $1\times10^{-4}$ during 200k iterations. All input images are resized to 518$\times$518 pixels for training and test.

\subsection{Datasets and Evaluation Metrics}
\paragraph{Datasets.}
For the performance evaluation of the proposed method, three human centric multi-view datasets (i.e., ZJU-Mocap~\cite{peng2021neural}, HuMMan~\cite{cai2022humman}, and THuman2.0~\cite{yu2021function4d}) are adopted.
Two real-world datasets, ZJU-Mocap and HuMMan, were captured by multiple calibrated cameras in controlled indoor environments.
For ZJU-Mocap, nine subjects are split into six for training and three for test, as follows by~\cite{hu2023sherf}.
For HuMMan, we follow the official split~\cite{cai2022humman}, which utilizes 317 sequences for training and 22 sequences for test.
We also evaluate our method on THuman2.0 dataset which provides 526 high-quality textured 3D human scans along with corresponding SMPL annotations. 
Following previous methods~\cite{rogsplat, ghg}, we construct a human-body prior by splitting the dataset into 426 subjects for training and 100 subjects for testing.
% ody prior by splitting the dataset into 426 subjects for training and 100 subjects for testing.

% \vspace{-1cm}
\paragraph{Evaluation metrics.}
The reconstruction performance is evaluated by using three metrics, i.e., Peak Signal-to-Noise Ratio (PSNR), Structural Similarity Index (SSIM)~\cite{SSIM}, and Learned Perceptual Image Patch Similarity (LPIPS)~\cite{LPIPS}.
PSNR and SSIM measure pixel-wise similarity and structural fidelity between the rendered image and the ground truth, while LPIPS reflects perceptual similarity in the feature domain.

%%%  이거 NERF-based  %%%%
\begin{table}[h]
\small
\centering
\fontsize{9.5pt}{11pt}\selectfont
\renewcommand{\arraystretch}{1.1}
\setlength{\tabcolsep}{4pt}
\begin{tabular}{l | c c c}
\toprule
\multicolumn{1}{l}{Methods} 
& PSNR($\uparrow$) & SSIM($\uparrow$) & LPIPS($\downarrow$) \\
\midrule
\toprule
SHERF~\cite{hu2023sherf}        
& 19.25 & 0.8942 & 0.1121 \\
NHP~\cite{kwon2021neural}            
& 25.74 & 0.9356 & 0.0748 \\
GP-NeRF~\cite{gao2022mps}     
& 23.28 & 0.9325 & 0.0798 \\
TransHuman~\cite{Trnashuman} 
& 27.36 & 0.9487 & 0.0505 \\
\textbf{Proposed method}  
& \textbf{30.93} & \textbf{0.9710} & \textbf{0.0334} \\
\bottomrule
\end{tabular}
\vspace{-1mm}
\caption{
\label{table:thuman_single}
Performance comparison with NeRF-based methods on the THuman2.0 dataset~\cite{yu2021function4d}.
The best results are shown in bold.
}
\end{table}
% \vspace{-7mm}
\subsection{Performance Evaluation}
% \vspace{-4mm}
\paragraph{Quantitative evaluation.}
To show the effectiveness of the proposed method, we compare ours with previous methods in rendering for human subjects with sparse input views, i.e., SHERF~\cite{hu2023sherf}, NHP~\cite{kwon2021neural}, GP-NeRF~\cite{gao2022mps}, TransHuman~\cite{Trnashuman}, GPS-Gaussian~\cite{gps-gaussian}, EVA-Gaussian~\cite{hu2024eva}, GHG~\cite{ghg}, and RoGSplat~\cite{rogsplat}.
First of all, the performance comparison on both ZJU-Mocap and HuMMan datasets is shown in Table~\ref{table:zju_humman}.
Compared with GHG~\cite{ghg} and RoGSplat~\cite{rogsplat}, our approach improves PSNR by a notable margin, particularly on the ZJU-Mocap dataset (+0.46dB over RoGSplat), indicating that the proposed method provides more accurate localization of 3D Gaussians.
Although the performance in the LPIPS metric is somewhat dropped, the proposed method consistently yields higher PSNR and SSIM scores, demonstrating sharper reconstruction and better structural preservation.
For the HuMMan dataset, which includes more complex motions and diverse human subjects captured under real-world conditions, our method also shows the competitive performance, demonstrating its robustness to both motion dynamics and subject variability.

% \vspace{1.9mm}
The performance comparison on the THuman2.0 dataset is shown in Table~\ref{table:thuman}. As can be seen, the proposed method show the meaningful improvement for the rendering performance. Specifically, the proposed method consistently achieves the highest PSNR, SSIM, and LPIPS scores regardless of the number of input views. 

% \vspace{1.9mm}
While NeRF-based approaches such as SHERF~\cite{hu2023sherf}, NHP~\cite{kwon2021neural}, GP-NeRF~\cite{gao2022mps}, and TransHuman~\cite{Trnashuman} rely on costly volumetric rendering, our Gaussian-based framework achieves the superior quality of human rendering with significantly reduced computational overhead.
In particular, the proposed method renders human subjects several times faster while maintaining high PSNR and perceptual fidelity.
This efficiency mainly comes from the explicit Gaussian representation and the feed-forward reconstruction pipeline, which eliminates the need for iterative ray sampling and per-scene optimization.
Consequently, we could confirm that the proposed method is suitable for real-time and large-scale scenarios of human rendering.
{
\setlength{\textfloatsep}{1pt}
\begin{figure}[!t]  % [t]: top 위치
    \centering
    \includegraphics[width=0.9\linewidth]{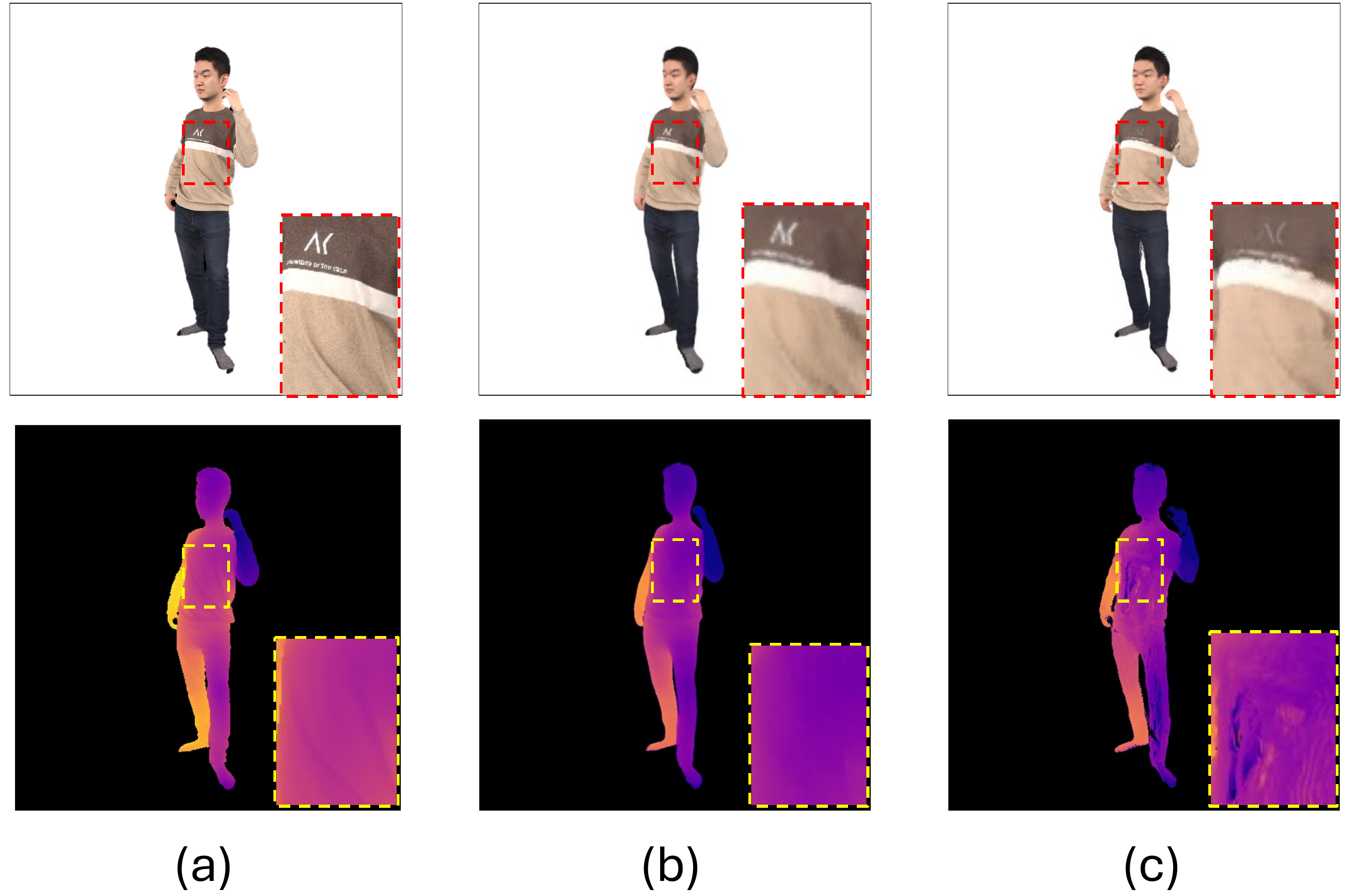}
    % \vspace{-3.0mm}
    \caption{
    Results of novel view synthesis (top row) and the corresponding depth map (bottom row) via generalizable human Gaussian splatting on the THuman2.0~\cite{yu2021function4d} dataset. (a) Ground truth. (b) Results by the proposed method. (c) Results by RoGSplat~\cite{rogsplat}.
}
    \label{fig:depth}
\end{figure}
% \vspace{-5mm}
}
% \FloatBarrier
% \vspace{-3mm}

\paragraph{Qualitative evaluation.}
The qualitative comparison of the proposed method with GHG~\cite{ghg} and RoGSplat~\cite{rogsplat} on ZJU-Mocap and HuMMan datasets, as shown in Fig.~\ref{fig:zju, humman}.
While existing methods often suffer from fine-grained clothing wrinkles (see the $2^{\text{nd}}$ row and $3^{\text{rd}}$ rows in  Figs.~\ref{fig:zju, humman}(b) and (c)), the complex facial region (see the $1^{\text{st}}$ and $2^{\text{nd}}$ rows in Figs.~\ref{fig:zju, humman}(b) and (c)), our method achieves superior fidelity, while preserving intricate human features and producing more realistic outputs.

% \vspace{-0.2mm}

We further compare the proposed method with GPS-Gaussian~\cite{gps-gaussian}, GHG~\cite{ghg}, and  RoGSplat~\cite{rogsplat} on the THuman2.0 dataset is presented in Fig.~\ref{fig:thuman}. 
As can be seen, the proposed method produces a clean surface without visible artifacts and better preserves local details, particularly around regions with complex geometry such as hair and loose clothing (see the first and second examples in Fig.~\ref{fig:thuman}).
In contrast, previous methods~\cite{rogsplat, ghg} relying on the SMPL-based Gaussian anchoring scheme often fail to represent non-rigid regions due to the limited expressive ability of the skinned mesh model.
By recalibrating latent embeddings in the 3D space through multi-view semantic consistency, the proposed method provides the reliable rendering result without significant textural distortions under sparse-view inputs.

% \vspace{-0.2mm}

To demonstrate the effectiveness of the proposed method in accurately localizing Gaussians, we compare rendering results and corresponding depth maps on the THuman2.0 dataset~\cite{yu2021function4d}, as shown in Fig.~\ref{fig:depth}. Whereas RoGSplat~\cite{rogsplat} yields inaccurate position of Gaussians, which leads to blurred textures in the rendering result (see the $3^{\text{rd}}$ column in Fig.~\ref{fig:depth}), the proposed method produces accurate depth values which help to preserve fine-grained details (see the $2^{\text{nd}}$ column in Fig.~\ref{fig:depth}).

\begin{table}[t]
\centering
\small
\setlength{\tabcolsep}{4pt}
\renewcommand{\arraystretch}{1.0}
\resizebox{1.0\linewidth}{!}{%
\begin{tabular}{ccccc}
\toprule
\multirow{2}{*}{3D Unproj.} &
\multirow{2}{*}{Semantic Cons.} &
\multicolumn{3}{c}{THuman2.0} \\
\cmidrule(lr){3-5}
& & PSNR ($\uparrow$) &
      SSIM ($\uparrow$) &
      LPIPS ($\downarrow$) \\
\midrule
\xmark & \cmark & 30.19 & 0.9676 & 0.0381 \\
\cmark & \xmark & 30.58 & 0.9695 & 0.0351 \\
\cmark & \cmark & \textbf{30.93} & \textbf{0.9710} & \textbf{0.0334} \\
\bottomrule
\end{tabular}
}
\caption{
\label{table:ablation}
Performance analysis of the proposed method according to different combinations of key components
in the proposed method on the THuman2.0 dataset.
}
\end{table}
% \vspace{-6mm}
\begin{figure}[h]  % [t]: top 위치
    \centering
    \includegraphics[width=\linewidth]{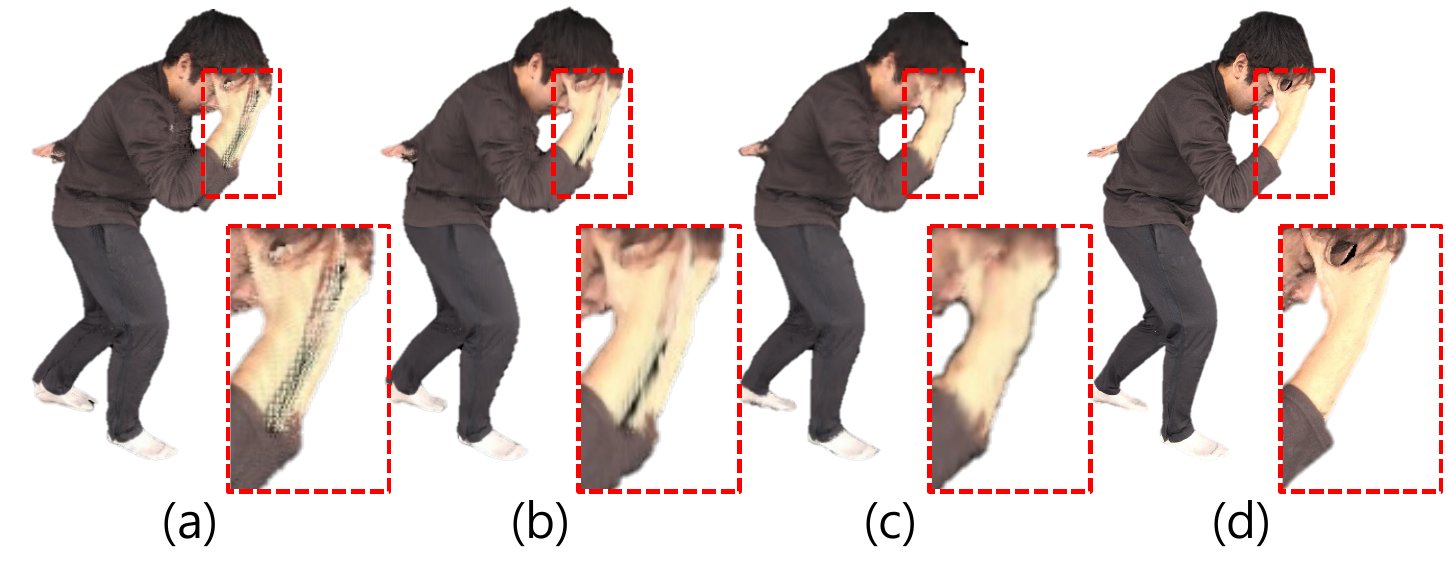}
    % \captionsetup{skip=2pt}
    % \vspace{-2mm}
    \caption{
Results of novel-view rendering on the THuman2.0 dataset. (a) Rendering without 3D unprojection. (b) Rendering without semantic features. (c) Proposed method. (d) Ground truth. }
    \label{fig:ablation}
\end{figure}
%% view 
% \vspace{-1.5mm}
\subsection{Ablation Study}
% \vspace{-1.5mm}
\quad To analyze the effectiveness of each component in the proposed method, we conduct ablation experiments on the THuman2.0 dataset under the 4-view setting.
Specifically, we evaluate the contribution of unprojecting latent embeddings and weighting the semantic consistency by progressively removing them from the proposed method.
The quantitative results are summarized in Table~\ref{table:ablation}, and qualitative comparisons are also shown in Fig.~\ref{fig:ablation}.
When the unprojection step is removed, the mismatched correspondence between multi-view inputs is hardly resolved while yielding errors in depth prediction, which makes Gaussian positions misaligned around occluded parts (see Fig.~\ref{fig:ablation}(a)).
As a result, the rendering performance is accordingly dropped.
The proposed method effectively resolves this issue by unprojecting latent embeddings into the 3D space, which is shared by all the input views.
In this 3D space, we can compute the spatial proximity between points of latent embeddings more accurately and further determine whether those belong to the same body part or not.
This alignment leads to more stable Gaussian localization and eliminates floating artifacts in complex body configurations.
In the following, we also check the effect of weighting the semantic consistency in computing the score of cross-view attention.
To do this, the semantic consistency term $s_{j,k}$ is set to $1$ in Eq~\ref{eq:semantic_attention}. Without considering the semantic consistency, the model tends to be confused between points of latent embeddings, which are from different body parts but closely positioned.
This causes blurred boundaries and loss of details, and further degrades the rendering quality. Note that the semantic consistency can be computed in the 3D space after the unprojection step, thus the case that only utilizes the semantic consistency is not included in Table 4.
Based on these two components, the proposed method achieves the best performance while providing sharp silhouettes and clean surfaces for the reliable rendering result of the target subject.
The visual comparison according to the change of components in the proposed method is also shown in Fig.~\ref{fig:fig3}.

\begin{figure}[t]  % [t]: top 위치
    \centering
    \includegraphics[width=0.90\linewidth]{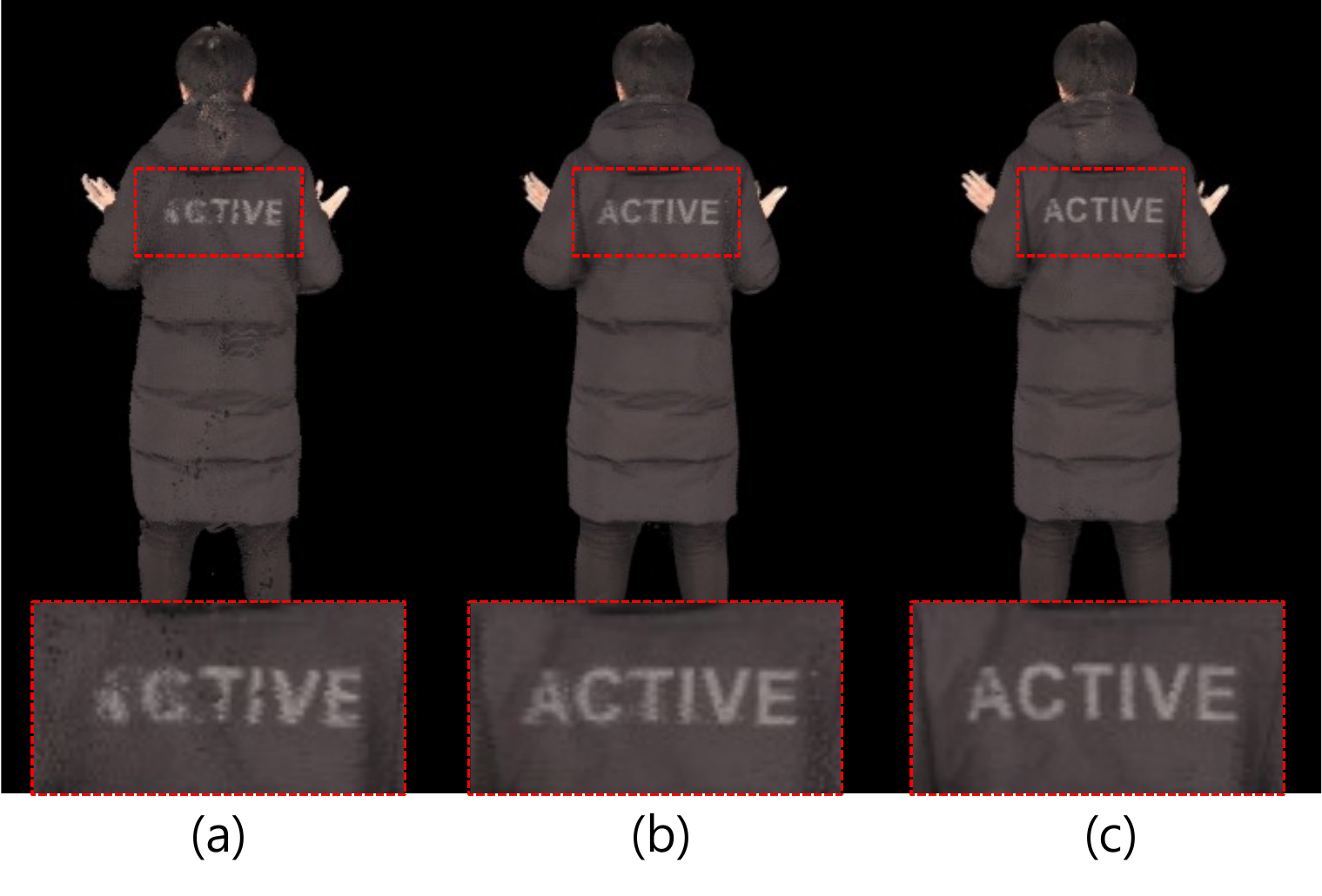}
    \vspace{-3mm}
    \caption{
Novel-view rendering results on the THuman2.0 dataset.
(a) Rendering result with 3-view inputs. (b) Rendering result with 4-view inputs. (c) Rendering result with 5-view inputs. }
    \label{fig:Fig8}
\end{figure}

To further demonstrate the robustness of our method to the number of input views, we perform an additional study on the THuman2.0 dataset by gradually reducing the number of available cameras from 5 to 3. As can be seen in Fig.~\ref{fig:Fig8}, the proposed method yields reliable novel-view rendering results when the number of input views is reduced (see Fig.~\ref{fig:Fig8}(a) and (b)), which demonstrates that the proposed method is robust to sparse-view settings.

% \vspace{-1.5mm}
\section{Conclusion}
% \vspace{-1.5mm}
\quad In this paper, we present a novel generalizable human Gaussian splatting method for sparse-view inputs.
The key idea of the proposed method is to accurately localize the position of Gaussians by leveraging semantic consistency across multi-view inputs.
To this end, our approach lifts multi-view image features into a unified 3D space by utilizing depth cues inferred from the input views.
Subsequently, the lifted features are recalibrated by a cross-view attention mechanism, which enforces the semantic consistency belonging to the same body part across views.
This is notably effective to estimate the accurate position of 3D Gaussians since it resolves spatial ambiguities in complex textures and occluded regions.
Extensive experiments on benchmark datasets demonstrate that our method consistently improves reconstruction quality and achieves state-of-the-art performance in generalizable human Gaussian splatting.
% \vspace{-3mm}
\paragraph{Acknowledgements.}
% This work was supported by the National Research Foundation of Korea (NRF) grant funded by the Korea government (MSIT) (RS-2023-NR076462) and Institute of Information Communications Technology Planning Evaluation (IITP) grant funded by the Korea government (MSIT) (No. 2018-0-00207, RS-2018-II180207, Immersive Media Research Laboratory).
This work was supported by Institute of Information \& Communications Technology Planning \& Evaluation (IITP) grant funded by the Korea government through the Ministry of Science and ICT (MSIT) (No. 2018-0-00207, RS-2018-II180207, Immersive Media Research Laboratory).
% This work was supported by Institute of Information and Communications Technology Planning and Evaluation (IITP) grant funded by the Korea government through the Ministry of Science and ICT (MSIT) (No. 2018-0-00207, RS-2018-II180207, Immersive Media Research Laboratory).
{
    \small
    \bibliographystyle{ieeenat_fullname}
    \bibliography{main}
}

%     \small
%     \bibliographystyle{ieeenat_fullname}
%     \bibliography{main}
% }

\end{document}